\documentclass[11pt,a4paper]{article}
\usepackage[hyperref]{acl2018}
\usepackage{times}
\usepackage{latexsym}

\usepackage{url}

\usepackage{multirow}
\usepackage{graphicx}
\usepackage{subfig}

\usepackage[T1]{fontenc}

\usepackage{CJKutf8}
\usepackage{array}
\usepackage{enumitem}
\usepackage{amssymb}

\aclfinalcopy 
\def\aclpaperid{396} 

\title{Analogical Reasoning on Chinese Morphological and Semantic Relations}

\makeatletter
\def\thanks#1{\protected@xdef\@thanks{\@thanks
        \protect\footnotetext{#1}}}
\makeatother

\author{
  Shen Li\textsuperscript{1,2,$^{\spadesuit}$} \And
  Zhe Zhao\textsuperscript{3,$^{\clubsuit}$} \And
  Renfen Hu\textsuperscript{1,2,$^{\spadesuit}$,$^{\dag}$\thanks{$^{\dag}$ Corresponding author.}{}} \And
  Wensi Li\textsuperscript{1,2,$^{\blacklozenge}$} \And
  Tao Liu\textsuperscript{3,$^{\clubsuit}$} \And
  Xiaoyong Du\textsuperscript{3,$^{\clubsuit}$} \AND
  $^{\spadesuit}${\tt \{shen, irishere\}@mail.bnu.edu.cn} \\
  $^{\clubsuit}${\tt \{helloworld, tliu, duyong\}@ruc.edu.cn} \\
  $^{\blacklozenge}${\tt zjklws@163.com} \\
  \textsuperscript{1} Institute of Chinese Information Processing, Beijing Normal University \\
  \textsuperscript{2} UltraPower-BNU Joint Laboratory for Artificial Intelligence, Beijing Normal University \\
  \textsuperscript{3} School of Information, Renmin University of China \\
}

\date{}

\begin{document}
\maketitle
\begin{CJK}{UTF8}{gbsn}

\begin{abstract}
Analogical reasoning is effective in capturing linguistic regularities. This paper proposes an analogical reasoning task on Chinese. After delving into Chinese lexical knowledge, we sketch 68 implicit morphological relations and 28 explicit semantic relations. A big and balanced dataset CA8 is then built for this task, including 17813 questions. Furthermore, we systematically explore the influences of vector representations, context features, and corpora on analogical reasoning. With the experiments, CA8 is proved to be a reliable benchmark for evaluating Chinese word embeddings.

\end{abstract}

\section{Introduction}

Recently, the boom of word embedding draws our attention to analogical reasoning on linguistic regularities. Given the word representations, analogy questions can be automatically solved via vector computation, e.g. \emph{``apples - apple + car $\approx$ cars''} for morphological regularities and \emph{``king - man + woman $\approx$ queen''} for semantic regularities \cite{mikolov2013linguistic}. Analogical reasoning has become a reliable evaluation method for word embeddings. In addition, It can be used in inducing morphological transformations \cite{soricut2015unsupervised}, detecting semantic relations \cite{herdadelenaa2009bagpack}, and translating unknown words \cite{langlais2007translating}.

It is well known that linguistic regularities vary a lot among different languages. For example, Chinese is a typical analytic language which lacks inflection. Figure 1 shows that function words and reduplication are used to denote grammatical and semantic information. In addition, many semantic relations are closely related with social and cultural factors, e.g. in Chinese \emph{``shī-xiān''} (god of poetry) refers to the poet \emph{Li-bai} and \emph{``shī-shèng''} (saint of poetry) refers to the poet \emph{Du-fu}.

\begin{figure}[t]
\begin{small}
\setlength{\belowcaptionskip}{-0.5cm}
\includegraphics[width=3in]{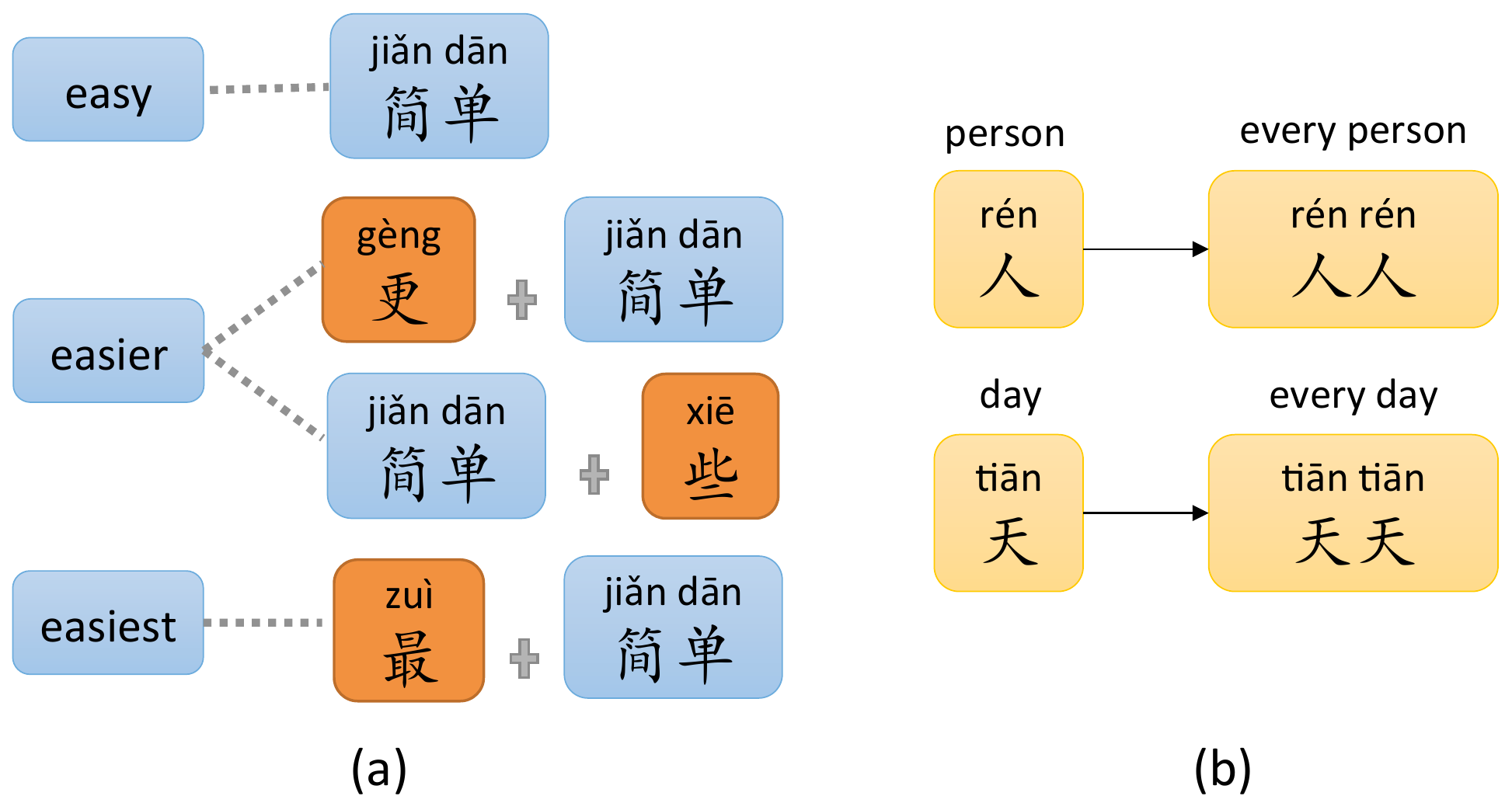}
\vspace*{-0.5em}
\caption{Examples of Chinese lexical knowledge:
(a) function words (in orange boxes) are used to indicate the comparative and superlative degrees;
(b) reduplication yields the meaning of \emph{``every''}.}
\end{small}
\vspace*{-0.5em}
\end{figure}

However, few attempts have been made in Chinese analogical reasoning. The only Chinese analogy dataset is translated from part of an English dataset \cite{chen2015joint} (denote as CA\_translated). Although it has been widely used in evaluation of word embeddings \cite{yang2015improved,yin2016multi,su2017learning}, it could not serve as a reliable benchmark since it includes only 134 unique Chinese words in three semantic relations (capital, state, and family), and morphological knowledge is not even considered.

Therefore, we would like to investigate linguistic regularities beneath Chinese. By modeling them as an analogical reasoning task, we could further examine the effects of vector offset methods in detecting Chinese morphological and semantic relations. As far as we know, this is the first study focusing on Chinese analogical reasoning. Moreover, we release a standard benchmark for evaluation of Chinese word embedding, together with 36 open-source pre-trained embeddings at GitHub\footnote{\href{https://github.com/Embedding/Chinese-Word-Vectors}{https://github.com/Embedding/Chinese-Word-Vectors}}, which could serve as a solid basis for Chinese NLP tasks.

\section{Morphological Relations}
Morphology concerns the internal structure of words. There is a common belief that Chinese is a morphologically impoverished language since a morpheme mostly corresponds to an orthographic character, and it lacks apparent distinctions between roots and affixes. However, \newcite{packard2000morphology} suggests that Chinese has a different morphological system because it selects different ``settings'' on parameters shared by all languages. We will clarify this special system by mapping its morphological analogies into two processes: reduplication and semi-affixation.

\subsection{Reduplication}
Reduplication means a morpheme is repeated to form a new word, which is semantically and/or syntactically distinct from the original morpheme, e.g. the word \emph{``tiān-tiān''(day day)} in Figure 1(b) means \emph{``everyday''}. By analyzing all the word categories in Chinese, we find that nouns, verbs, adjectives, adverbs, and measure words have reduplication abilities. Given distinct morphemes A and B, we summarize 6 repetition patterns in Figure 2.

\begin{figure}[bthp]
\begin{small}
\includegraphics[width=3in]{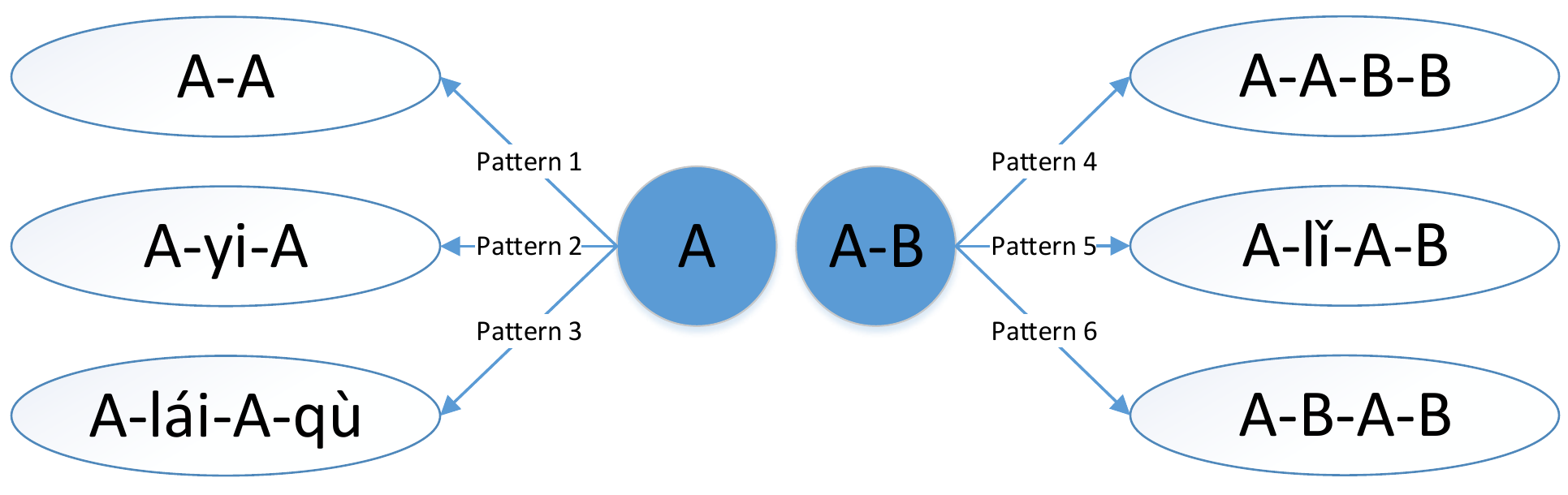}
\caption{Reduplication patterns of A and A-B.}
\end{small}
\end{figure}

Each pattern may have one or more morphological functions. Taking \emph{Pattern 1 (A$\rightarrow$AA)} as an example, noun morphemes could form kinship terms or yield every/each meaning. For verbs, it signals doing something a little bit or things happen briefly. AA reduplication could also intensify an adjective or transform it to an adverb.

\begin{itemize}
  \vspace*{-0.5em}
  \itemsep-0.2em
  \item \emph{\underline{bà}(dad) $\rightarrow$ \underline{bà-bà}(dad)}
  \item \emph{\underline{tiān}(day) $\rightarrow$ \underline{tiān-tiān}(everyday)}
  \item \emph{\underline{shuō}(say) $\rightarrow$ \underline{shuō-shuo}(say a little)}
  \item \emph{\underline{kàn}(look) $\rightarrow$ \underline{kàn-kàn}(have a brief look)}
  \item \emph{\underline{dà}(big) $\rightarrow$ \underline{dà-dà}(very big; greatly)}
  \item \emph{\underline{shēn}(deep) $\rightarrow$ \underline{shēn-shēn}(deeply)}
\end{itemize}

\subsection{Semi-affixation}
Affixation is a morphological process whereby a bound morpheme (an affix) is attached to roots or stems to form new language units. Chinese is a typical isolating language that has few affixes. \newcite{liu2001pratical} points out that although affixes are rare in Chinese, there are some components behaving like affixes and can also be used as independent lexemes. They are called semi-affixes.

To model the semi-affixation process, we uncover 21 semi-prefixes and 41 semi-suffixes. These semi-suffixes can be used to denote changes of meaning or part of speech. For example, the semi-prefix \emph{``dì-''} could be added to numerals to form ordinal numbers, and the semi-suffix \emph{``-zi''} is able to nominalize an adjective:

\begin{itemize}
  \vspace*{-0.5em}
  \itemsep-0.2em
  \item \emph{\underline{yī}(one) $\rightarrow$ \underline{dì-yī}(first)\\\underline{èr}(two) $\rightarrow$ \underline{dì-èr}(second)}
  \item \emph{\underline{pàng}(fat) $\rightarrow$ \underline{pàng-zi}(a fat man)\\\underline{shòu}(thin) $\rightarrow$ \underline{shòu-zi}(a thin man)}
\end{itemize}

\section{Semantic Relations}
To investigate semantic knowledge reasoning, we present 28 semantic relations in four aspects: geography, history, nature, and people. Among them we inherit a few relations from English datasets, e.g. country-capital and family members, while the rest of them are proposed originally on the basis of our observation of Chinese lexical knowledge. For example, a Chinese province may have its own abbreviation, capital city, and representative drama, which could form rich semantic analogies:

\begin{itemize}
  \vspace*{-0.5em}
  \itemsep-0.2em  
  \item \underline{\emph{ān-huī}}\space\space vs\space\space\underline{\emph{zhè-jiāng}} (province)
  \item \underline{\emph{wǎn}}\space\space vs\space\space\underline{\emph{zhè}} (abbreviation)
  \item \underline{\emph{hé-féi}}\space\space vs\space\space\underline{\emph{háng-zhōu}} (capital)
  \item \underline{\emph{huáng-méi-xì}}\space\space vs\space\space\underline{\emph{yuè-jù}} (drama)
\end{itemize}

We also address novel relations that could be used for other languages, e.g. scientists and their findings, companies and their founders.

\section{Task of Chinese Analogical Reasoning}

\begin{table*}[!htb]\label{benchmark}
\begin{small}
  \centering
\newcommand{\tabincell}[2]{\begin{tabular}{@{}#1@{}}#2\end{tabular}}
    \begin{tabular}{|c|c|c|c|c|c|}

    \hline
    Benchmark&Category&Type&\#questions&\#words&Relation\\
    \hline
\multirow{3}{*}{CA\_translated} &\multirow{3}{*}{Semantic}
            &Capital& 506 & 46 &capital-country \\
            \cline{3-6}
      & &State& 175 & 54 &city-province \\
            \cline{3-6}
            & &Family& 272 & 34 &family members \\
      \hline
            \hline
\multirow{8}{*}{CA8} &\multirow{4}{*}{Morphological}
            &Reduplication A& 2554 & 344 &A-A, A-yi-A, A-lái-A-qù \\
            \cline{3-6}
      & &Reduplication AB& 2535 & 423 &A-A-B-B, A-lǐ-A-B, A-B-A-B \\
            \cline{3-6}
            & &Semi-prefix& 2553 & 656 &21 semi-prefixes: 大, 小, 老, 第, 亚, etc.\\
            \cline{3-6}
            & &Semi-suffix& 2535 & 727 &41 semi-suffixes: 者, 式, 主义, 性, etc.\\
      \cline{2-6}
&\multirow{4}{*}{Semantic}
            &Geography & 3192 & 305 & \tabincell{c}{country-capital, country-currency,\\province-abbreviation, province-capital,\\province-dramma, etc. }  \\
            \cline{3-6}
      & &History& 1465 & 177 & \tabincell{c}{dynasty-emperor, dynasty-capital,\\title-emperor, celebrity-country} \\
            \cline{3-6}
            & &Nature& 1370 & 452 & \tabincell{c}{number, time, animal, plant, body,\\physics, weather, reverse, color, etc.} \\
            \cline{3-6}
            & &People& 1609 & 259 &\tabincell{c}{finding-scientist, work-writer,\\family members, etc.} \\
      \hline
    \end{tabular}
\end{small}
\caption{Comparisons of CA\_translated and CA8 benchmarks. More details about the relations in CA8 can be seen in \href{https://github.com/Embedding/Chinese-Word-Vectors}{GitHub}. }
\end{table*}

\begin{table*}
\begin{small}
\newcommand{\tabincell}[2]{\begin{tabular}{@{}#1@{}}#2\end{tabular}}
  \centering
    \begin{tabular}{|c|c|c|c|c|c|>{\hspace{-0.4em}}c<{\hspace{-0.4em}}|c|}
      \hline
\tabincell{c}{Window\\(dynamic)}&Iteration&Dimension&\tabincell{c}{Sub-\\sampling}&\tabincell{c}{Low-frequency\\threshold}&\tabincell{c}{Context distribution\\ smoothing}&\tabincell{c}{Negative\\ (SGNS/PPMI)}&\tabincell{c}{Vector\\ offset}\\
            \hline
5&5&300&1e-5&50&0.75&5/1&3COSMUL\\
      \hline
\end{tabular}
\end{small}
\caption{Hyper-parameter details. \newcite{levy2014neural} unifies SGNS and PPMI in a framework, which share the same hyper-parameter settings. We exploit 3COSMUL to solve the analogical questions suggested by \newcite{levy2014linguistic}.}
\end{table*}

Analogical reasoning task is to retrieve the answer of the question ``a is to b as c is to ?''. Based on the relations discussed above, we firstly collect word pairs for each relation. Since there are no explicit word boundaries in Chinese, we take dictionaries and word segmentation specifications as references to confirm the inclusion of each word pair. To avoid the imbalance problem addressed in English benchmarks \cite{gladkova2016analogy}, we set a limit of 50 word pairs at most for each relation. In this step, 1852 unique Chinese word pairs are retrieved.
We then build CA8, a big, balanced dataset for Chinese analogical reasoning including 17813 questions. Compared with CA\_translated \cite{chen2015joint}, CA8 incorporates both morphological and semantic questions, and it brings in much more words, relation types and questions. Table 1 shows details of the two datasets. They are both used for evaluation in Experiments section.

\section{Experiments}

In Chinese analogical reasoning task, we aim at investigating to what extent word vectors capture the linguistic relations, and how it is affected by three important factors: vector representations (sparse and dense), context features (character, word, and ngram), and training corpora (size and domain). Table 2 shows the hyper-parameters used in this work. All the text data used in our experiments (as shown in Table 3) are preprocessed via the following steps:
\begin{itemize}
    \vspace{-0.5em}
    \itemsep-0.2em
    \item Remove the html and xml tags from the texts and set the encoding as utf-8. Digits and punctuations are remained.
    \item Convert traditional Chinese characters into simplified characters with Open Chinese Convert (OpenCC)\footnote{\href{https://github.com/BYVoid/OpenCC}{https://github.com/BYVoid/OpenCC}}.
    \item Conduct Chinese word segmentation with HanLP(v\_1.5.3)\footnote{\href{https://github.com/hankcs/HanLP}{https://github.com/hankcs/HanLP}}.
\end{itemize}

\subsection{Vector Representations}

\newcommand{\specialcell}[2][c]{%
  \begin{tabular}[#1]{@{}c@{}}#2\end{tabular}}
\begin{table*}[!htb]
\begin{small}
\centering
\begin{tabular}{|c|c|c|c|c|}
\hline
Corpus&Size&\#tokens&$|V|$&Description\\
\hline
Wikipedia&1.3G&223M&2129K&\specialcell{Wikipedia data obtained from \\ \href{https://dumps.wikimedia.org/}{https://dumps.wikimedia.org/}}\\
\hline
Baidubaike&4.1G&745M&5422K&\specialcell{Chinese wikipedia data from \\ \href{https://baike.baidu.com/}{https://baike.baidu.com/}}\\
\hline
People's Daily News&3.9G&668M&1664K&\specialcell{News data from People's Daily (1946-2017)\\ \href{http://data.people.com.cn/}{http://data.people.com.cn/}}\\
\hline
Sogou news&3.7G&649M&1226K&\specialcell{News data provided by Sogou Labs \\ \href{http://www.sogou.com/labs/}{http://www.sogou.com/labs/}}\\
\hline
Zhihu QA&2.1G&384M&1117K&\specialcell{Chinese QA data from \href{https://www.zhihu.com/}{https://www.zhihu.com/},\\ including 32137 questions and 3239114 answers}\\
\hline
Combination&14.8G&2668M&8175K&We build this corpus by combining the above corpora\\
\hline
\end{tabular}
\caption{Detailed information of the corpora. \#tokens denotes the number of tokens in corpus. $|V|$ denotes the vocabulary size.}
\end{small}
\end{table*}

\begin{table*}
\begin{small}
  \centering
		\begin{tabular}{|>{\hspace{-0.2em}}cc<{\hspace{-0.2em}}|>{\hspace{-0.2em}}ccc<{\hspace{-0.2em}}||>{\hspace{-0.2em}}cccc<{\hspace{-0.2em}}||>{\hspace{-0.2em}}c<{\hspace{-0.2em}}||>{\hspace{-0.2em}}cccc<{\hspace{-0.2em}}||>{\hspace{-0.2em}}c<{\hspace{-0.2em}}|}
      \hline
      \multicolumn{2}{|c|}{\multirow{2}{*}{ }}
            & \multicolumn{3}{c||}{CA\_translated} & \multicolumn{10}{c|}{CA8} \\
            \cline{3-15}
			& & Cap. & Sta. & Fam. & A & AB & Pre. & Suf. & \textbf{Mor.} & Geo. & His. & Nat. & Peo. & \textbf{Sem.} \\
            \cline{3-15}
			\hline
\multirow{3}{*}{SGNS}
            &word&.706&.966&.603                 &.117&.162&.181&.389&.222   &.414&.345&.236&.223&.327 \\
            \cline{2-15}
			&word+ngram&.715&\textbf{.977}&.640      &.143&.184&.197&.429&.250    &.449&.308&.276&.310&.368 \\
            \cline{2-15}
            &word+char&.676&.966&.548               &\textbf{.358}&\textbf{.540}&\textbf{.326}&\textbf{.612}&\textbf{.455}          &.468&.226&.296&.305&.368 \\
            \cline{2-15}
			\hline
\multirow{3}{*}{PPMI}
            &word&.925&.920&.548                 &.103&.139&.138&.464&.226       &.627&.501&.300&.515&.522 \\
            \cline{2-15}
			&word+ngram&\textbf{.943}&.960&\textbf{.658}      &.102&.129&.168&.456&.230          &\textbf{.680}&\textbf{.535}&\textbf{.371}&\textbf{.626}&\textbf{.586} \\
            \cline{2-15}
            &word+char&.913&.886&.614                 &.106&.190&.173&.505&.260       &.638&.502&.288&.515&.524 \\
            \cline{2-15}
			\hline
		\end{tabular}
\end{small}
\caption{Performance of word representations learned under different configurations. Baidubaike is used as the training corpus. The top 1 results are in \textbf{bold}.}
\end{table*}

Existing vector representations fall into two types, dense vectors and sparse vectors. SGNS (skip-gram model with negative sampling) \cite{mikolov2013linguistic} and PPMI (Positive Pointwise Mutual Information) \cite{levy2014linguistic} are respectively typical methods for learning dense and sparse word vectors. Table 4 lists the performance of them on CA\_translated and CA8 datasets under different configurations.

We can observe that on CA8 dataset, SGNS representations perform better in analogical reasoning of morphological relations and PPMI representations show great advantages in semantic relations. This result is consistent with performance of English dense and sparse vectors on MSR (morphology-only), SemEval (semantic-only), and Google (mixed) analogy datasets \cite{levy2014neural,TACL570}. It is probably because the reasoning on morphological relations relies more on common words in context, and the training procedure of SGNS favors frequent word pairs. Meanwhile, PPMI model is more sensitive to infrequent and specific word pairs, which are beneficial to semantic relations.

The above observation shows that CA8 is a reliable benchmark for studying the effects of dense and sparse vectors. Compared with CA\_translated and existing English analogy datasets, it offers both morphological and semantic questions which are also balanced across different types \footnote{CA\_translated and SemEval datasets contain only semantic questions, MSR dataset contains only morphological questions, and in Google dataset the capital:country relation constitutes 56.72\% of all semantic questions.}.

\subsection{Context Features}

To investigate the influence of context features on analogical reasoning, we consider not only word features, but also ngram features inspired by statistical language models, and character (Hanzi) features based on the close relationship between Chinese words and their composing characters \footnote{The SGNS with word and character features are implemented by \href{https://github.com/facebookresearch/fastText}{fasttext toolkit}, the rest are implemented by \href{https://github.com/zhezhaoa/ngram2vec}{ngram2vec toolkit}. }. Specifically, we use word bigrams for ngram features, character unigrams and bigrams for character features.

Ngrams and Chinese characters are effective features in training word representations \cite{zhao2017ngram2vec,chen2015joint,bojanowski2016enriching}. However, Table 4 shows that there is only a slight increase on CA\_translated dataset with ngram features, and the accuracies in most cases decrease after integrating character features. In contrast, on CA8 dataset, the introduction of ngram and character features brings significant and consistent improvements on almost all the categories. Furthermore, character features are especially advantageous for reasoning of morphological relations. SGNS model integrating with character features even doubles the accuracy in morphological questions.

Besides, the representations achieve surprisingly high accuracies in some categories of CA\_translated, which means that there is little room for further improvement. However it is much harder for representation methods to achieve high accuracies on CA8. The best configuration only achieves 68.0\%.

\begin{table*}[!htb]
\begin{small}
  \centering
    \begin{tabular}{|>{\hspace{-0.5em}}c<{\hspace{-0.5em}}|ccc||cccc||c||cccc||c<{\hspace{-0.5em}}|}
    \hline
\multirow{2}{*}{}
            & \multicolumn{3}{c||}{CA\_translated} & \multicolumn{10}{c|}{CA8} \\
            \cline{2-14}
      & Cap. & Sta. & Fam. & A & AB & Pre. & Suf. &\textbf{Mor.} & Geo. & His. & Nat. & Peo. &\textbf{Sem.} \\
            \cline{2-14}
            \hline
            Wikipedia 1.2G           &.597&.771&.360        &.029&.018&.152&.266&.180                    &.339&.125&.147&.079&.236 \\
            \hline
            Baidubaike 4.3G         & .706& .966& .603        & .117& .162& .181&\textbf{.389}&.222                & .414&\textbf{.345}& .236&\textbf{.223}&.327 \\
            \hline
            People's Daily 4.2G  &\textbf{.925}&\textbf{.989}& .547         & .140& .158&\textbf{.213}& .355&\textbf{.226}        &\textbf{.694}& .019& .206& .157&\textbf{.455} \\
            \hline
            Sogou News 4.0G         & .619& .966& .496        & .057& .075& .131& .176&.115        & .432& .067& .150& .145&.302 \\
            \hline
            Zhihu QA 2.2G              & .277& .491&\textbf{.625}          &\textbf{.175}&\textbf{.199}& .134& .251&.189        & .146& .147&\textbf{.250}& .189&.181 \\
            \hline
            Combination 15.9G             &\textbf{.872}&\textbf{.994}&\textbf{.710}     &\textbf{.223}&\textbf{.300}&\textbf{.234}&\textbf{.518}&\textbf{.321}    &\textbf{.662}&\textbf{.293}&\textbf{.310}&\textbf{.307}&\textbf{.467} \\
      \hline
    \end{tabular}
\end{small}
\caption{Performance of word representations learned upon different training corpora by SGNS with context feature of word. The top 2 results are in \textbf{bold}.}
\end{table*}

\subsection{Corpora}
We compare word representations learned upon corpora of different sizes and domains. As shown in Table 3, six corpora are used in the experiments: Chinese Wikipedia, Baidubaike, People's Daily News, Sogou News, Zhihu QA, and ``Combination'' which is built by combining the first five corpora together.

Table 5 shows that accuracies increase with the growth in corpus size, e.g. Baidubaike (an online Chinese encyclopedia) has a clear advantage over Wikipedia. Also, the domain of a corpus plays an important role in the experiments. We can observe that vectors trained on news data are beneficial to geography relations, especially on People's Daily which has a focus on political news. Another example is Zhihu QA, an online question-answering corpus which contains more informal data than others. It is helpful to reduplication relations since many reduplication words appear frequently in spoken language. With the largest size and varied domains, ``Combination'' corpus performs much better than others in both morphological and semantic relations.

Based on the above experiments, we find that vector representations, context features, and corpora all have important influences on Chinese analogical reasoning. Also, CA8 is proved to be a reliable benchmark for evaluation of Chinese word embeddings.

\section{Conclusion}
In this paper, we investigate the linguistic regularities beneath Chinese, and propose a Chinese analogical reasoning task based on 68 morphological relations and 28 semantic relations. In the experiments, we apply vector offset method to this task, and examine the effects of vector representations, context features, and corpora. This study offers an interesting perspective combining linguistic analysis and representation models. The benchmark and embedding sets we release could also serve as a solid basis for Chinese NLP tasks.

\makeatletter
\def\@seccntformat#1{
  \expandafter\ifx\csname c@#1\endcsname\c@section\else
  \csname the#1\endcsname\quad
  \fi}
\makeatother
\section{Acknowledgments}
This work is supported by the Fundamental Research Funds for the Central Universities, National Natural Science Foundation of China with Grant (No.61472428) and Chinese Testing International Project (No.CTI2017B12).

\end{CJK}

\bibliography{acl2018}
\bibliographystyle{acl_natbib}

\end{document}